\documentclass{article}

\PassOptionsToPackage{numbers, compress}{natbib}

\usepackage[preprint]{neurips_2022}

\usepackage[utf8]{inputenc} 
\usepackage[T1]{fontenc}    
\usepackage{hyperref}       
\usepackage{url}            
\usepackage{booktabs}       
\usepackage{amsfonts}       
\usepackage{nicefrac}       
\usepackage{microtype}      
\usepackage{xcolor}         

\RequirePackage{algorithm}
\RequirePackage{algorithmic}

\usepackage{multirow}
\usepackage{amsmath}
\usepackage{capt-of}
\usepackage{tabularx}
\usepackage{epsfig}
\usepackage{amssymb}
\usepackage{amsfonts}
\usepackage{booktabs}
\usepackage{scalerel}
\usepackage[inline]{enumitem}
\usepackage{listings}
\usepackage{varwidth}
\usepackage[export]{adjustbox}
\usepackage{tikz}
\usetikzlibrary{tikzmark}

\usepackage{stmaryrd}
\usepackage{bbm}
\usepackage{wrapfig}
\usepackage{pifont}

\newcommand{\tabincell}[2]{\begin{tabular}{@{}#1@{}}#2\end{tabular}}

\newcommand{\sptk}[1]{\texttt{[#1]}}

\definecolor{deepblue}{rgb}{0,0,0.5}
\definecolor{officeblue}{RGB}{0,102,204}
\definecolor{deepred}{rgb}{0.6,0,0}
\definecolor{deepgreen}{rgb}{0,0.5,0}
\definecolor{mybrickred}{RGB}{182,50,28}

\definecolor{fillcolor}{RGB}{216,217,252}

\usepackage{amsmath,amsfonts,bm}

\def\eqref#1{equation~\ref{#1}}

\def\1{\bm{1}}

\def\vh{{\bm{h}}}

\def\vv{{\bm{v}}}
\def\vw{{\bm{w}}}

\def\mH{{\bm{H}}}

\def\mT{{\bm{T}}}

\def\mV{{\bm{V}}}

\DeclareMathAlphabet{\mathsfit}{\encodingdefault}{\sfdefault}{m}{sl}
\SetMathAlphabet{\mathsfit}{bold}{\encodingdefault}{\sfdefault}{bx}{n}

\newcommand{\R}{\mathbb{R}}

\newcommand{\softmax}{\mathrm{softmax}}

\usepackage{pifont}
\newcommand{\cmark}{{\color{blue}\ding{51}}}%
\newcommand{\xmark}{{\color{red}\ding{55}}}%

\newcommand\our{\textsc{VLMo}}
\newcommand\mome{\textsc{MoME}}
\newcommand\beit{\textsc{BEiT}}
\newcommand{\tblidx}[1]{{\small \texttt{[#1]}}}

\title{\our{}: Unified Vision-Language Pre-Training with Mixture-of-Modality-Experts}

\author{{Hangbo Bao\thanks{~Equal contribution. $\dagger$ Contact person.}, Wenhui Wang\footnotemark[1], Li Dong, Qiang Liu} \\ 
\textbf{Owais Khan Mohammed, Kriti Aggarwal, Subhojit Som, Furu Wei$^\dagger$} \\
Microsoft \\
\url{https://aka.ms/vlmo}
}

\begin{document}

\maketitle

\begin{abstract}
We present a unified \textbf{V}ision-\textbf{L}anguage pretrained~\textbf{Mo}del (\textbf{\our{}}) that jointly learns a dual encoder and a fusion encoder with a modular Transformer network. Specifically, we introduce \textbf{M}ixture-\textbf{o}f-\textbf{M}odality-\textbf{E}xperts (\textbf{\mome{}}) Transformer, where each block contains a pool of modality-specific experts and a shared self-attention layer. Because of the modeling flexibility of \mome{}, pretrained \our{} can be fine-tuned as a fusion encoder for vision-language classification tasks, or used as a dual encoder for efficient image-text retrieval. Moreover, we propose a stagewise pre-training strategy, which effectively leverages large-scale image-only and text-only data besides image-text pairs. Experimental results show that \our{} achieves state-of-the-art results on various vision-language tasks, including VQA, NLVR2 and image-text retrieval. The code and pretrained models are available at \url{https://aka.ms/vlmo}.
\end{abstract}

\section{Introduction}
\label{sec:intro}

Vision-Language (VL) pre-training~\cite{vilbert,vl-bert,clip,oscar,vilt,albef} learns generic cross-modal representations from large-scale image-text pairs.
Previous models usually employ image-text matching, image-text contrastive learning, masked region classification/feature regression, word-region/patch alignment and masked language modeling to aggregate and align visual and linguistic information.
Then the pretrained models can be directly fine-tuned on downstream vision-language tasks, such as VL retrieval and classification (visual question answering, visual reasoning, etc.).

Two mainstream architectures are widely used in previous work. 
CLIP~\citep{clip} and ALIGN~\citep{align} adopt a \textit{dual-encoder} architecture to encode images and text separately. 
Modality interaction is handled by the cosine similarity of the image and text feature vectors.
The dual-encoder architecture is effective for retrieval tasks, especially for masses of images and text.
Feature vectors of images and text can be pre-computed and stored.
However, the shallow interaction between images and text is not enough to handle complex VL classification tasks.
ViLT~\citep{vilt} finds that CLIP gives a relatively low accuracy on visual reasoning task.
Another line of work~\citep{vilbert,vl-bert,lxmert,uniter,vilt,albef} relies on a fusion encoder with cross-modal attention to model image-text pairs.
Multi-layer Transformer~\citep{transformer} networks are usually employed to fuse image and text representations.
The \textit{fusion-encoder} architecture achieves superior performance on VL classification tasks.
But it requires to jointly encode all possible image-text pairs to compute similarity scores for retrieval tasks.
The quadratic time complexity leads to a much slower inference speed than the dual-encoder models whose time complexity is linear.

In order to take advantage of the two types of architectures, we propose a unified \textbf{V}ision-\textbf{L}anguage pretrained \textbf{Mo}del (\textbf{\our{}}) that can be used as either a dual encoder to separately encode images and text for retrieval tasks, or used as a fusion encoder to model the deep interaction of image-text pairs for classification tasks.
This is achieved by introducing \textbf{M}ixture-\textbf{o}f-\textbf{M}odality-\textbf{E}xperts (\textbf{\mome{}}) Transformer that can encode various modalities (images, text, and image-text pairs) within a Transformer block.
\mome{} employs a pool of modality experts to replace the feed-forward network in standard Transformer.
It captures modality-specific information by switching to different modality experts, and uses the shared self-attention across modalities to align visual and linguistic information.
Specifically, \mome{} Transformer consists of three modality experts, namely vision expert for image encoding, language expert for text encoding, and vision-language expert for image-text fusion.
Thanks to the modeling flexibility, we can reuse \mome{} Transformer with the shared parameters for different purposes, i.e., text-only encoder, image-only encoder, and image-text fusion encoder.

\our{} is jointly learned with three pre-training tasks, namely image-text contrastive learning, image-text matching, and masked language modeling.
In addition, we propose a stagewise pre-training strategy to effectively leverage large-scale image-only and text-only corpus besides image-text pairs in \our{} pre-training.
We first pretrain vision experts and self-attention modules of \mome{} Transformer on image-only data using masked image modeling proposed in \beit{}~\citep{beit}. 
We then pretrain language experts on text-only data using masked language modeling~\citep{bert}.
Finally, the model is used to initialize vision-language pre-training.
By getting rid of the limited size of image-text pairs and their simple and short captions, stagewise pre-training on large amounts of image-only and text-only data helps \our{} to learn more generalizable representations.

Experimental results demonstrate that \our{} achieves state-of-the-art results on vision-language retrieval and classification tasks.
Our model, used as a dual encoder, outperforms fusion-encoder-based models~\citep{uniter,villa,vilt,albef} while enjoying a much faster inference speed on retrieval tasks.
Moreover, our model also achieves state-of-the-art results on visual question answering (VQA) and natural language for visual reasoning (NLVR2), where \our{} is used as a fusion encoder.

Our main contributions are summarized as follows:
\begin{itemize}[leftmargin=1.5em]
\item We propose a unified vision-language pretrained model \our{} that can be used as a fusion encoder for classification tasks, or fine-tuned as a dual encoder for retrieval tasks.
\item We introduce a general-purpose multimodal Transformer for vision-language tasks, namely \mome{} Transformer, to encode different modalities. It captures modality-specific information by modality experts, and aligns contents of different modalities by the self-attention module shared across modalities.
\item We show that stagewise pre-training using large amounts of image-only and text-only data greatly improves our vision-language pretrained model.
\end{itemize}

\begin{figure*}[t]
\begin{center}
\begin{tabular}{c}
\includegraphics[width=0.98\textwidth]{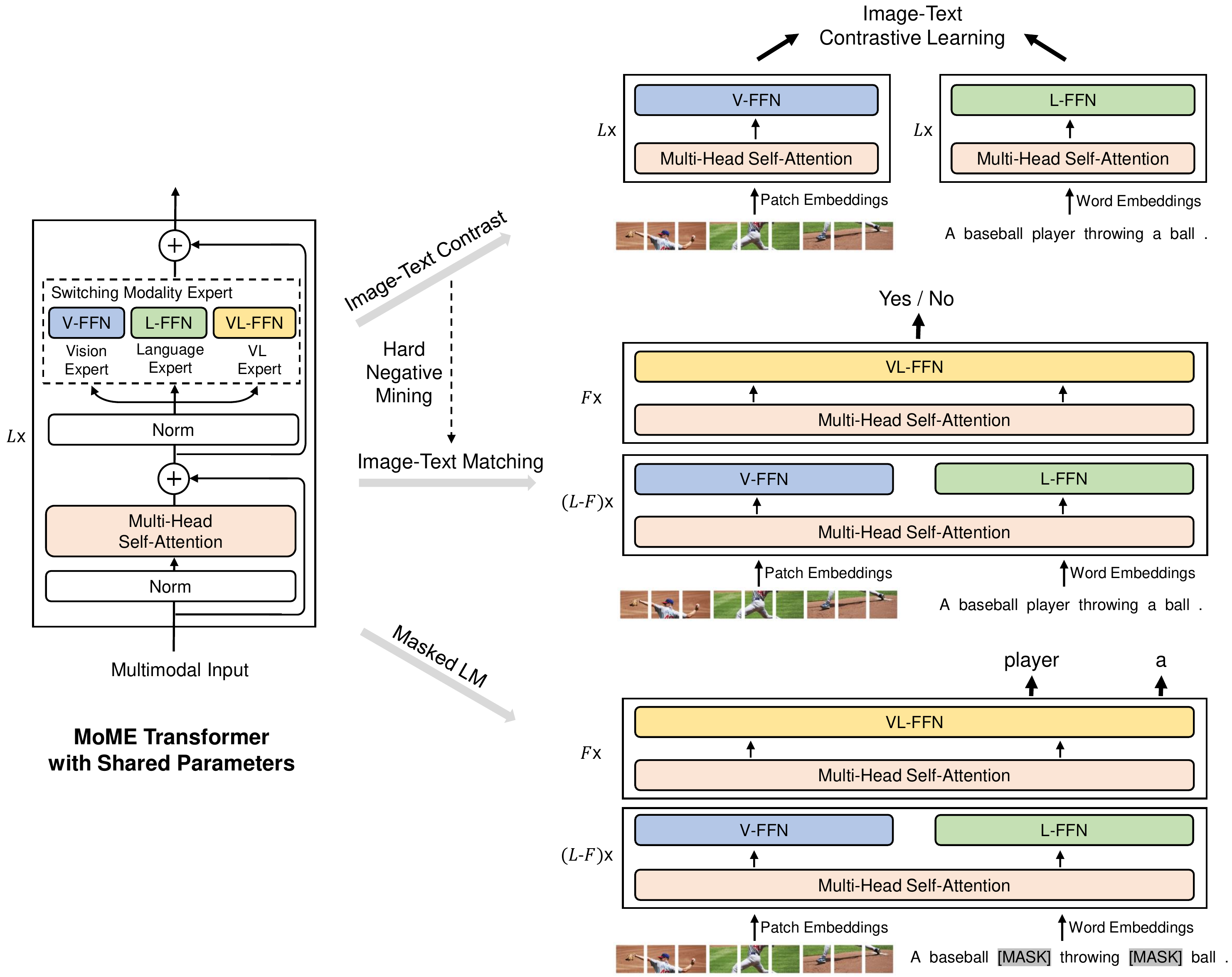}
\end{tabular}
\end{center}
\caption{Overview of \our{} pre-training. 
We introduce mixture-of-modality-experts (\mome{}) Transformer to encode different modality input by modality-specific experts.
The model parameters are shared across image-text contrastive learning, masked language modeling, and image-text matching pre-training tasks.
During fine-tuning, the flexible modeling enables us to use \our{} as either a dual encoder (i.e., separately encode images and text for retrieval tasks) or a fusion encoder (i.e., jointly encode image-text pairs for better interaction across modalities).
}
\label{fig:overview}
\end{figure*}

\section{Related Work}
\label{sec:related:work}

Pre-training with Transformer~\citep{transformer} backbone networks has substantially advanced the state of the art across natural language processing~\citep{gpt,bert,roberta,bart,unilm,t5,unilm2,xlm,xlmr,xnlg,infoxlm,xlme,deltalm}, computer vision~\citep{vit,deit,beit} and vision-language~\citep{lxmert,vl-bert,uniter,vinvl,clip,align,vilt,albef} tasks.

The approaches of vision-language pre-training can be divided into two categories.
The first category utilizes a dual encoder to encode images and text separately, and uses cosine similarity or a linear projection layer to model the interaction between images and text~\citep{clip,align}.
Image-text contrastive learning is usually employed to optimize the model.
Dual-encoder models are effective for vision-language retrieval tasks.
However, the simple interaction is not enough to handle tasks that require complex reasoning, such as visual reasoning and visual question answering (VL classification tasks). 
The second category models the interaction of images and text using a deep fusion encoder with cross-modal attention~\citep{lxmert,vilbert,vl-bert,visualbert,unifiedvlp,uniter,oscar,unimo,villa,vinvl,pixel-bert,soho,vilt,albef,simvlm}. 
Image-text matching, masked language modeling, word-region/patch alignment, masked region classification and feature regression are widely used to train fusion-encoder-based models.
These models achieve better performance for vision-language classification tasks, while the joint encoding of all image-text pairs leads to a slow inference speed for retrieval tasks.
A large portion of fusion-encoder-based models rely on an off-the-shelf object detector like Faster R-CNN~\citep{faster-rcnn} to obtain image region features. 
Generating region features slows down the inference speed and renders the approach less scalable.
Recently, Pixel-BERT~\citep{pixel-bert} removes object detector and encodes images into grid features by convolutional neural networks.
ALBEF~\citep{albef} employs image Transformer~\citep{vit,deit} to obtain the representations of images, and uses text Transformer~\citep{bert} to learn the contextualized representations of text.
These representations are then fused by cross-modal attention.
ViLT~\citep{vilt} encodes images into patch embeddings, and then feed the concatenation of image patch embeddings and word embeddings into a Transformer network to learn contextualized representations and model the interaction of images and text.

Different from previous work, 
our unified pre-training using shared \mome{} Transformer enables the model perform separate encoding for retrieval tasks, and jointly encode image-text pairs to capture deeper interaction for classification tasks.
Our model achieves competitive performance, while enjoying a faster inference speed for both retrieval and classification tasks.

\section{Methods}
\label{sec:methods}

Given image-text pairs, \our{} obtains image-only, text-only and image-text pair representations by the \mome{} Transformer network.
As shown in Figure~\ref{fig:overview}, the unified pre-training optimizes shared \mome{} Transformer with image-text contrastive learning on image-only and text-only representations, image-text matching and masked language modeling on image-text pair representations.
Thanks to the modeling flexibility, the model can be used as a dual encoder for retrieval tasks to encode images and text separately during fine-tuning.
It can also be fine-tuned as a fusion encoder to model deeper modality interaction of images and text for classification tasks.

\subsection{Input Representations}
\label{sec:repr}

Given an image-text pair, we encode the pair into image, text and image-text vector representations.
These representations are then fed into the \mome{} Transformer to learn contextualized representations and align image and text feature vectors.

\paragraph{Image Representations} 
Following vision Transformers~\citep{vit,deit,beit}, the 2D image $\vv \in \R^{H \times W \times C}$ is split and reshaped into $N={HW}/{P^2}$ patches $\vv^{p} \in \R^{N \times (P^2 C)}$, where $C$ is the number of channels, $(H, W)$ is the resolution of the input image, and $(P, P)$ is the patch resolution. 
The image patches are then flattened into vectors and are linearly projected to obtain patch embeddings.
We also prepend a learnable special token \sptk{I\_CLS} to the sequence.
Finally, image input representations are obtained via summing patch embeddings, learnable 1D position embeddings $\mV_{pos} \in \R^{(N+1) \times D}$ and image type embedding $\mV_{type} \in \R^{D}$: $\mH_0^{v} = [ \vv_{\sptk{I\_CLS}} , \mV \vv^{p}_{i} , \dots , \mV \vv^{p}_{N} ] + \mV_{pos} + \mV_{type}$, 
where $\mH_0^{v} \in \R^{(N+1) \times D}$, linear projection $\mV \in \R^{(P^2 C) \times D}$.

\paragraph{Text Representations}
Following BERT~\citep{bert}, we tokenize the text to subword units by WordPiece~\citep{gnmt}. 
A start-of-sequence token (\sptk{T\_CLS}) and a special boundary token (\sptk{T\_SEP}) are added to the text sequence. 
Text input representations $\mH_0^{w} \in \R^{(M+2) \times D}$ are computed via summing the corresponding word embedding, text position embedding and text type embedding $\mH_0^{w} = [ \vw_{\sptk{T\_CLS}} , \vw_{i} , \dots , \vw_{M} , \vw_{\sptk{T\_SEP}} ] + \mT_{pos} + \mT_{type}$. 
$M$ indicates the length of tokenized subword units.

\paragraph{Image-Text Representations}
We concatenate image and text input vectors to form the image-text input representations $\mH_0^{vl} = [\mH_0^{w} ; \mH_0^{v}]$

\begin{figure*}[t]
\begin{center}
\begin{tabular}{c}
\includegraphics[width=0.96\textwidth]{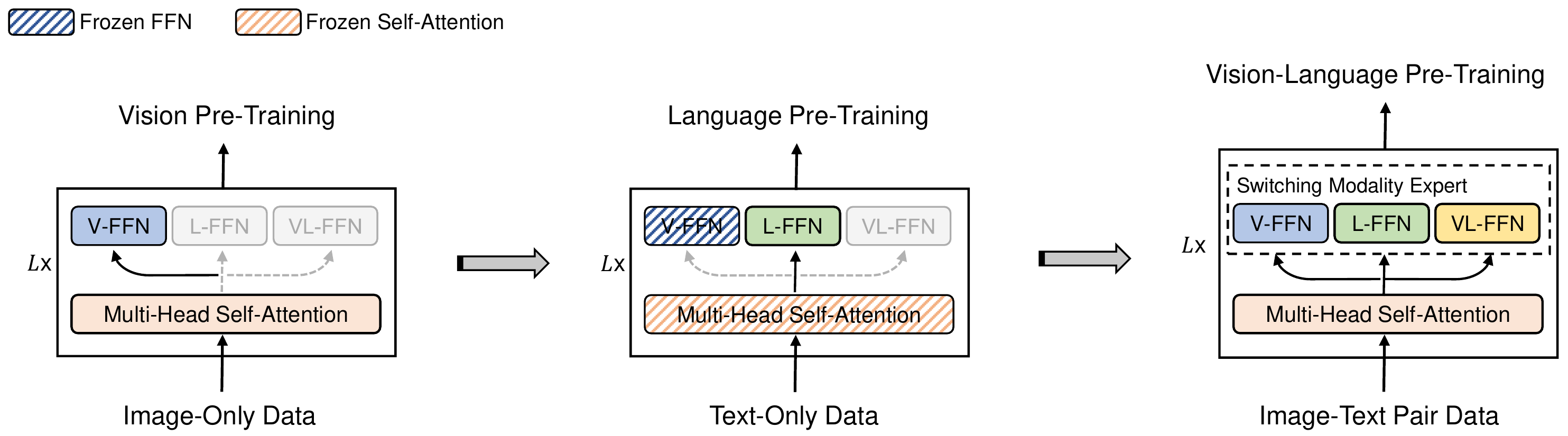}
\end{tabular}
\end{center}
\caption{
Stagewise pre-training using image-only and text-only corpora.
We first pretrain the vision expert (V-FFN) and self-attention module on large-scale image-only data as in \beit{}~\citep{beit}.
Then the parameters of vision expert and self-attention module are frozen, and we train the language expert (L-FFN) by masked language modeling on large amounts of text-only data.
Finally, we train the whole model with vision-language pre-training.
}
\label{fig:image_text_training}
\end{figure*}

\subsection{Mixture-of-Modality-Experts Transformer}
\label{sec:mome}

Inspired by mixture-of-experts networks~\citep{moe,switchtransformer}, we propose a general-purpose multimodal Transformer for vision-language tasks, namely \mome{} Transformer, to encode different modalities.
\mome{} Transformer introduces mixture of modality experts as a substitute of the feed forward network of standard Transformer.
Given previous layer's output vectors $\mH_{l-1}, l \in [1, L]$, each \mome{} Transformer block captures modality-specific information by switching to different modality expert, and employs multi-head self-attention (MSA) shared across modalities to align visual and linguistic contents. LN is short for layer normalization.
\begin{align}
\mH_{l}^{\prime} = \mathrm{MSA}(\mathrm{LN}(\mH_{l-1}))+\mH_{l-1} \\
\mH_{l} = \mathrm{MoME}\text{-}\mathrm{FFN}(\mathrm{LN}(\mH_{l}^{\prime}))+\mH_{l}^{\prime}
\end{align}
$\mathrm{MoME}\text{-}\mathrm{FFN}$ selects an expert among multiple modality experts to process the input according to the modality of the input vectors $\mH_{l}^{\prime}$ and the index of the Transformer layer.
Specifically, there are three modality experts: vision expert (V-FFN), language expert (L-FFN) and vision-language expert (VL-FFN).
If the input is image-only or text-only vectors, we use vision expert for encoding images and language expert for encoding text.
If the input consists of vectors of multiple modalities, such as the vectors of image-text pair, we employ vision expert and language expert to encode the respective modality vectors at the bottom Transformer layers.
Vision-language expert is then used at the top layers to capture more modality interaction.
Given the three types of input vectors, we obtain image-only, text-only and image-text contextualized representations.

\subsection{Pre-Training Tasks}

\our{} is jointly pretrained by image-text contrastive learning on the image and text representations, masked language modeling and image-text matching on the image-text pair representations with shared parameters. 

\paragraph{Image-Text Contrast}

Given a batch of $N$ image-text pairs, image-text contrastive learning aims to predict the matched pairs from $N \times N$ possible image-text pairs.
There are $N^{2} - N$ negative image-text pairs within a training batch.

The final output vectors of \sptk{I\_CLS} token and \sptk{T\_CLS} token are used as the aggregated representation of the image and text, respectively.
Followed by a linear projection and normalization, we obtain image vectors $\{ \hat{\vh}^{v}_{i} \}_{i=1}^{N}$ and text vectors $\{ \hat{\vh}^{w}_{i} \}_{i=1}^{N}$ in a training batch to compute image-to-text and text-to-image similarities:
\begin{gather}
s_{i,j}^{i2t} = \hat{\vh}^{v\intercal}_{i}\hat{\vh}^{w}_{j}, ~
s_{i,j}^{t2i} = \hat{\vh}^{w\intercal}_{i}\hat{\vh}^{v}_{j}
\\
p^{i2t}_{i} = \frac{\mathrm{exp}(s_{i,i}^{i2t} / \sigma)}{\sum_{j=1}^{N}{\mathrm{exp}(s_{i,j}^{i2t} / \sigma)}}, ~
p^{t2i}_{i} = \frac{\mathrm{exp}(s_{i,i}^{t2i} / \sigma)}{\sum_{j=1}^{N}{\mathrm{exp}(s_{i,j}^{t2i} / \sigma)}}
\end{gather}

Where $s_{i,j}^{i2t}$ represents image-to-text similarity of image of $i$-th pair and text of $j$-th pair,
$s_{i,j}^{t2i}$ is the text-to-image similarity.
$\hat{\vh}^{w}_{i} \in \R^{D}$ and $\hat{\vh}^{v}_{j}  \in \R^{D}$ indicate the normalized vectors of $i$-th text and $j$-th image,
$\sigma$ is a learned temperature parameter.
$p^{i2t}_{i}$ and $p^{t2i}_{i}$ are the $\softmax$-normalized similarities.
Cross-entropy losses over image-to-text and text-to-image similarities are used to train the model.  

\begin{figure*}[t]
\begin{center}
\begin{tabular}{c}
\includegraphics[width=0.96\textwidth]{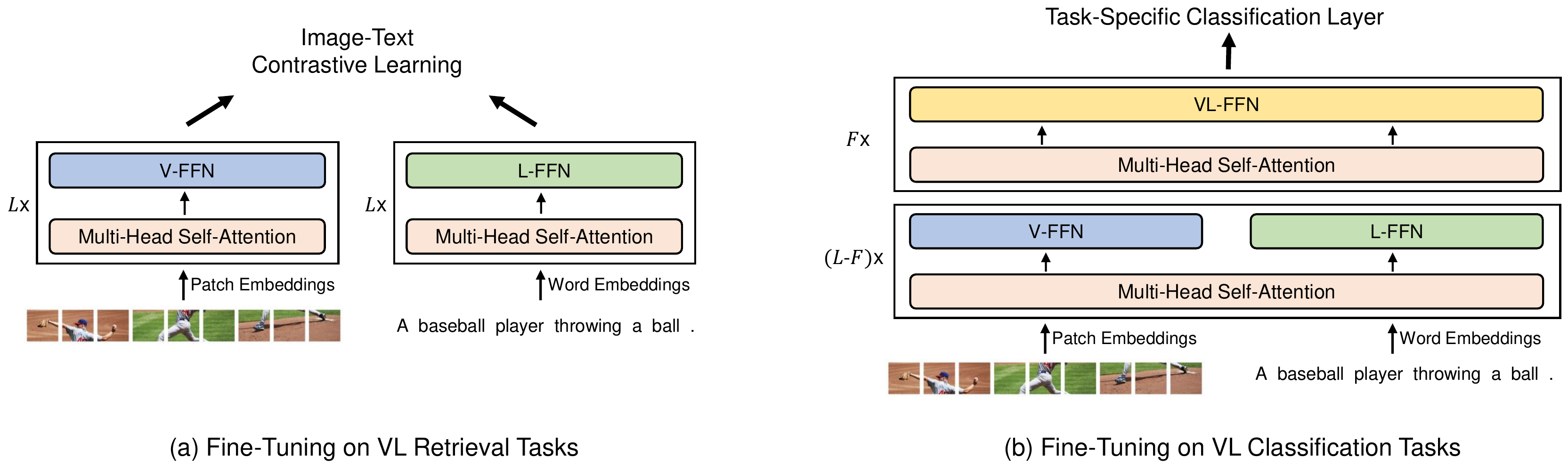}
\end{tabular}
\end{center}
\caption{
Fine-tuning \our{} on vision-language retrieval and classification tasks.
The model can be fine-tuned as a dual encoder to separately encode image and text for retrieval tasks.
\our{} can also be used as a fusion encoder to handle interaction of image-text pairs for classification tasks.
}
\label{fig:vlmo_finetuning}
\end{figure*}

\paragraph{Masked Language Modeling}

Following BERT~\citep{bert}, we randomly choose tokens in the text sequence, and replace them with the \sptk{MASK} token. 
The model is trained to predict these masked tokens from all the other unmasked tokens and vision clues. 
We use $15$\% masking probability as in BERT.
The final output vectors of masked tokens are fed into a classifier over the whole text vocabulary with cross-entropy loss.

\paragraph{Image-Text Matching}

Image-text matching aims to predict whether the image and text is matched.
We use the final hidden vector of the \sptk{T\_CLS} token to represent the image-text pair, and feed the vector into a classifier with cross-entropy loss for binary classification.
Inspired by ALBEF~\citep{albef}, we sample hard negative image-text pairs based on the contrastive image-to-text and text-to-image similarities.
Different from ALBEF~\citep{albef}, which samples hard negatives from training examples of the single GPU (we named it as local hard negative mining). 
We propose global hard negative mining and sample hard negative image-text pairs from more training examples gathered from all GPUs.
Global hard negative mining can find more informative image-text pairs and significantly improves our model.

\subsection{Stagewise Pre-Training}
\label{sec:stagewise_pretraining}

We introduce a stagewise pre-training strategy, which leverages large-scale image-only and text-only corpus to improve the vision-language model.
As present in Figure~\ref{fig:image_text_training}, we first perform vision pre-training on image-only data, and then perform language pre-training on text-only data to learn general image and text representations. 
The model is used to initialize the vision-language pre-training to learn the alignment of visual and linguistic information.
For vision pre-training, we train the attention module and vision expert of \mome{} Transformer as in \beit{}~\citep{beit} on image-only data.
We directly utilize the pretrained parameters of \beit{} to initialize the attention module and vision expert.
For language pre-training, we freeze parameters of the attention module and vision expert, and utilize masked language modeling~\citep{bert} to optimize the language expert on text-only data.
Compared with image-text pairs, image-only and text-only data are easier to collect.
In addition, text data of image-text pairs is usually short and simple.
Pre-training on image-only and text-only corpus improves the generalization on complex pairs.

\subsection{Fine-Tuning \our{} on Downstream Tasks}
\label{sec:ft}

As present in Figure~\ref{fig:vlmo_finetuning}, our model can be fine-tuned to adapt to various vision-language retrieval and classification tasks. 

\paragraph{Vision-Language Classification}
For classification tasks such as visual question answering and visual reasoning, \our{} is used as a fusion encoder to model modality interaction of images and text.
We use the final encoding vector of the token \sptk{T\_CLS} as the representation of the image-text pair, and feed it to a task-specific classifier layer to predict the label.

\paragraph{Vision-Language Retrieval}
For retrieval tasks, \our{} can be used as a dual encoder to encode images and text separately.
During fine-tuning, our model is optimized for the image-text contrastive loss. 
During inference, we compute representations of all images and text, and then use dot product to obtain image-to-text and text-to-image similarity scores of all possible image-text pairs. 
Separate encoding enables a much faster inference speed than fusion-encoder-based models.

\begin{table*}[t]
\centering
\small
\begin{tabular}{lccccc}
\toprule
\multirow{2}{*}{\bf Model} & \multirow{2}{*}{\textbf{\tabincell{c}{\# Pretrain \\ Images}}} & \multicolumn{2}{c}{\bf VQA} & \multicolumn{2}{c}{\bf NLVR2} \\
 & & test-dev & test-std & dev & test-P \\
\midrule
\multicolumn{5}{l}{\textit{ Base-Size Models Pretrained on COCO, VG, SBU and CC datasets}} \\
UNITER-Base~\citep{uniter} & 4M & 72.70 & 72.91 & 77.18 & 77.85 \\
VILLA-Base~\citep{villa} & 4M & 73.59 & 73.67 & 78.39 & 79.30 \\
UNIMO-Base~\citep{unimo} & 4M & 73.79 & 74.02 & - & - \\
ViLT-Base~\citep{vilt} & 4M & 71.26 & - & 75.70 & 76.13 \\
ALBEF-Base~\citep{albef} & 4M & 74.54 & 74.70 & 80.24 & 80.50 \\
\bf \our{}-Base & 4M & \bf 76.64 & \bf 76.89 & \bf 82.77 & \bf 83.34 \\
\midrule
\multicolumn{5}{l}{\textit{ Large-Size Models Pretrained on COCO, VG, SBU and CC datasets}} \\
UNITER-Large~\citep{uniter} & 4M & 73.82 & 74.02 & 79.12 & 79.98 \\
VILLA-Large~\citep{villa} & 4M & 74.69 & 74.87 & 79.76 & 81.47 \\
UNIMO-Large~\citep{unimo} & 4M & 75.06 & 75.27 & - & - \\
\bf \our{}-Large & 4M & \bf 79.94 & \bf 79.98 & \bf 85.64 & \bf 86.86 \\
\midrule
\multicolumn{5}{l}{\textit{ Models Pretrained on More Data}} \rule{0pt}{2.5ex} \\
VinVL-Large~\citep{vinvl} & 5.7M & 76.52 & 76.60 & 82.67 & 83.98 \\
SimVLM-Large~\citep{simvlm} & 1.8B & 79.32 & 79.56 & 84.13 & 84.84 \\
SimVLM-Huge~\citep{simvlm} & 1.8B & 80.03 & 80.34 & 84.53 & 85.15 \\
Florence-Huge~\citep{florence} & 900M & 80.16 & 80.36 & - & - \\
\bf \our{}-Large++ & 1.0B & \bf 82.88 & \bf 82.78 & \bf 88.62 & \bf 89.54 \\
\bottomrule
\end{tabular}
\caption{Fine-tuning results of base-size and large-size \our{} on vision-language classification datasets.
\our{}-Large++ is the model trained on one billion noisy image-text pairs with a larger batch size.
We report vqa-score on VQA test-dev and test-standard split, and report accuracy for NLVR2 development and public test set (test-P).
}
\label{tbl:results:classification}
\end{table*}

\section{Experiments}
\label{sec:exp}

We pretrain our model using large-scale image-text pairs and evaluate the model on visual-linguistic classification and retrieval tasks.

\subsection{Pre-Training Setup}
\label{sec:pretraining:setup}

Following previous work~\citep{uniter,vilt}, our pre-training data consists of four image captioning datasets: Conceptual Captions (CC)~\citep{gcc}, SBU Captions~\citep{sbu}, COCO~\citep{coco} and Visual Genome (VG)~\citep{vg} datasets.
There are about $4$M images and $10$M image-text pairs in the pre-training data.

\begin{table*}[t]
\centering
\small
\begin{tabular}{@{\hskip1pt}l@{\hskip1pt} @{\hskip1pt}c@{\hskip1pt} @{\hskip1pt}c@{ \hskip1pt} @{\hskip1pt}c@{ \hskip1pt} @{\hskip1pt}c@{ \hskip1pt} @{\hskip1pt}c@{ \hskip1pt} @{\hskip1pt}c@{ \hskip1pt} @{\hskip1pt}c@{ \hskip1pt} | @{ \hskip2pt}c@{ \hskip1pt} @{\hskip1pt}c@{ \hskip1pt} @{\hskip1pt}c@{ \hskip1pt} @{\hskip1pt}c@{ \hskip1pt} @{\hskip1pt}c@{ \hskip1pt} @{\hskip1pt}c@{ \hskip1pt} }
\toprule
\multirow{3}{*}{\bf Model} & \multirow{3}{*}{\textbf{\tabincell{c}{\# Pretrain \\ Images}}} & \multicolumn{6}{c}{\bf MSCOCO (5K test set)} & \multicolumn{6}{c}{\bf Flickr30K (1K test set)} \\
 & & \multicolumn{3}{c}{\bf TR} & \multicolumn{3}{c}{\bf IR} & \multicolumn{3}{c}{\bf TR} & \multicolumn{3}{c}{\bf IR} \\
 & & R@1 & R@5 & R@10 & R@1 & R@5 & R@10 & R@1 & R@5 & R@10 & R@1 & R@5 & R@10 \\
\midrule
\multicolumn{14}{l}{\textit{ Base-Size Models Pretrained on COCO, VG, SBU and CC datasets}} \\
UNITER-Base & 4M &  64.4 & 87.4 & 93.1 & 50.3 & 78.5 & 87.2 & 85.9 & 97.1 & 98.8 & 72.5 & 92.4 & 96.1 \\
VILLA-Base & 4M & - & - & - & - & - & - & 86.6 & 97.9 & 99.2 & 74.7 & 92.9 & 95.8 \\
ViLT-Base & 4M & 61.5 & 86.3 & 92.7 & 42.7 & 72.9 & 83.1 & 83.5 & 96.7 & 98.6 & 64.4 & 88.7 & 93.8 \\
ALBEF-Base$\ddag$ & 4M & 73.1 & 91.4 & 96.0 & 56.8 & 81.5 & 89.2 & \bf 94.3 & \bf 99.4 & 99.8 & \bf 82.8 & \bf 96.7 & \bf 98.4 \\
\bf \our{}-Base$\dag$ & 4M & \bf 74.8 & \bf 93.1 & \bf 96.9 & \bf 57.2 & \bf 82.6 & \bf 89.8 & 92.3 & \bf 99.4 & \bf 99.9 & 79.3 & 95.7 & 97.8 \\
\midrule
\multicolumn{14}{l}{\textit{ Large-Size Models Pretrained on COCO, VG, SBU and CC datasets}} \\
UNITER-Large & 4M & 65.7 & 88.6 & 93.8 & 52.9 & 79.9 & 88.0 & 87.3 & 98.0 & 99.2 & 75.6 & 94.1 & 96.8 \\
VILLA-Large & 4M & - & - & - & - & - & - & 87.9 & 97.5 & 98.8 & 76.3 & 94.2 & 96.8 \\
\bf \our{}-Large$\dag$ & 4M & \bf 78.2 & \bf 94.4 & \bf 97.4 & \bf 60.6 & \bf 84.4 & \bf 91.0 & \bf 95.3 & \bf 99.9 & \bf 100.0 & \bf 84.5 & \bf 97.3 & \bf 98.6 \\
\midrule
\multicolumn{14}{l}{\textit{ Models Pretrained on More Data}} \rule{0pt}{2.5ex} \\
VinVL-Large & 5.7M & 75.4 & 92.9 & 96.2 & 58.8 & 83.5 & 90.3 & - & - & - & - & - & - \\
ALIGN-Large$\dag$ & 1.8B & 77.0 & 93.5 & 96.9 & 59.9 & 83.3 & 89.8 & 95.3 & 99.8 & \bf 100.0 & 84.9 & 97.4 & 98.6 \\
Florence-Huge$\dag$ & 900M & 81.8 & 95.2 & - & 63.2 & 85.7 & - & \bf 97.2 & 99.9 & - & 87.9 & 98.1 & - \\
\bf \our{}-Large++$\dag$ & 1.0B & \bf 83.1 & \bf 96.0 & \bf 98.2 & \bf 65.2 & \bf 86.5 & \bf 92.2 & 96.8 & \bf 100.0 & \bf 100.0 & \bf 88.1 & \bf 98.4 & \bf 99.3 \\
\bottomrule
\end{tabular}
\caption{Fine-tuning results of text-retrieval (TR) and
image-retrieval (IR) on COCO and Flickr30K.
$\dag$: ALIGN, Florence and our model encode images and text separately, and then employ a shallow interaction (dot product) to obtain the similarity scores.
$\ddag$: ALBEF first encodes images and text separately to obtain the top-$k$ candidates, and then feed these representations into a fusion encoder to rerank the candidates.
The others require to encode all image-text combinations by a fusion encoder.
\our{}-Large++ represents the model trained on one billion noisy image-text pairs with a larger batch size.
}
\label{tbl:results:retrieval}
\end{table*}

Our models adopt the same network configuration as ViT~\citep{vit} and \beit{}~\citep{beit}.
\our{}-Base consists of $12$-layer Transformer blocks with $768$ hidden size and $12$ attention heads.
\our{}-Large is a $24$-layer Transformer network with $1024$ hidden size and $16$ attention heads.
The intermediate size of feed-forward networks is $3072$ and $4096$ for base-size and large-size models, respectively. 
\our{}-Base uses vision-language expert on the top two Transformer layers, and \our{}-Large introduces vision-language expert on the top three layers. 
For images, the input resolution is $224 \times 224$ and the patch size is $16 \times 16$ during pre-training.
We apply RandAugment~\citep{randaugment} to the input images.
The tokenizer of the uncased version of BERT is employed to tokenize the text.
The maximum text sequence length is set to $40$.
We also employ whole word masking for the masked language modeling pre-training task. 
We pretrain the models for $200$k steps with $1024$ batch size.
We utilize AdamW~\citep{adamw} optimizer with $\beta_1=0.9$, $\beta_2=0.98$. 
The peak learning is 2e-4 for the base-size model, 5e-5 for the large-size model.
Weight decay is set to $0.01$.
We use linear warmup over the first $2.5$k steps and linear decay.
The vision-language pre-training of base-size model takes about two days using 64 Nvidia Tesla V100 32GB GPU cards, and the large-size model takes about three days using 128 Nvidia Tesla V100 32GB GPU cards.

\subsection{Training on Larger-scale Datasets}
\label{sec:pretraining:scaleup}
We scale up vision-language representation learning by training \our{}-Large on one billion noisy web image-text pairs with a larger batch size.
We first pretrain the model for 200k steps with 16k batch size, and then continue train the model for 100k steps with 32k batch size.
The other hyper-parameters are the same as the training on 4M data.
Please refer to the supplementary material for more details of hyper-parameters used for pre-training and fine-tuning.

\subsection{Evaluation on Vision-Language Classification Tasks}
\label{sec:exp:classification}

We first conduct fine-tuning experiments on two widely used classification datasets: visual question answering~\citep{vqa} and natural language for visual reasoning~\citep{nlvr2}.
The model is fine-tuned as a fusion encoder to model deeper interaction.

\paragraph{Visual Question Answering (VQA)} For VQA, a natural image and a question are given, the task is to generate/choose the correct answer. 
We train and evaluate the model on VQA 2.0 dataset~\citep{vqa}.
Following common practices, we convert VQA 2.0 to a classification task, and choose the answer from a shared set consists of $3,129$ answers.
We use the final encoding vector of the \sptk{T\_CLS} token as the representation of the image-question pair and feed it to a classifier layer to predict the answer.

\paragraph{Natural Language for Visual Reasoning (NLVR2)}

The NLVR2~\citep{nlvr2} dataset requires the model to predict whether a text description is true about a pair of images.
Following OSCAR~\citep{oscar} and VinVL~\citep{vinvl}, we convert the triplet input to two image-text pairs, each containing the text description and one image.
We concatenate the final output vectors of the \sptk{T\_CLS} token of the two input pairs.
The concatenated vector is then fed into a classification layer to predict the label.

We present the results of VL classification tasks in Table~\ref{tbl:results:classification}. 
\our{} achieves state-of-the-art performance and substantially outperforms previous methods.
Our large-size model even outperforms SimVLM-Huge~\citep{simvlm} and Florence-Huge~\citep{florence} by a large margin, which consists of more parameters and are also trained on larger-scale image-text pairs.
Our model uses a simple linear projection to embed images as in ViLT~\citep{vilt}.
This leads to a significant speedup compared with previous models using image region features, which are extracted by an off-the-shelf object detector~\citep{vilbert,vl-bert,uniter,villa,unimo,vinvl}. 

\subsection{Evaluation on Vision-Language Retrieval Tasks}
\label{sec:exp:retrieval}

The retrieval tasks contain image-to-text retrieval and text-to-image retrieval. 
We evaluate the model on the widely used COCO~\citep{coco} and Flickr30K~\citep{flickr30k} datasets, and use the Karpathy split~\citep{karpathysplit} for both datasets.
The model is used as a dual encoder for retrieval tasks.
We encode images and text separately and compute their similarity scores by the dot product of image and text vectors.

As present in Table~\ref{tbl:results:retrieval}, \our{} achieves competitive performance with previous fusion-encoder-based models while having a much faster speed. 
Fusion-encoder-based models need to jointly encode all possible image-text pairs to compute their similarity scores, which requires quadratic time complexity.
Moreover, our large-size model even outperforms the huge-size model of Florence~\citep{florence}, which also trained on massive image-text pairs using a larger batch size.
\our{} pre-training can effectively leverage larger-scale noisy pairs and benefit from large batch training.

\subsection{Evaluation on Vision Tasks}

As shown in Table~\ref{tbl:unimodal_tasks}, we use \our{} as an image-only encoder and evaluate it on image classification (ImageNet~\cite{imagenet}) and semantic segmentation (ADE20K~\cite{ade20k}) tasks.
The model also achieves competitive performance, even slightly better than the \textsc{BEiT} model used for the initialization of \our{}.
The image resolution is 224$\times$224 for ImageNet, and 512$\times$512 for ADE20K.
We perform intermediate fine-tuning~\citep{beit} on ImageNet-21k for all three models.

\subsection{Ablation Studies}
\label{sec:ablation}

\begin{table*}[ht]
\centering
\small
\begin{tabular}{lcc}
\toprule
\bf Models & \bf ImageNet (acc@1) & \bf ADE20K (mIoU) \\ \midrule
\textsc{ViT}-Base & 83.6 & - \\ 
\textsc{BEiT}-Base & 85.2 & 52.8 \\ 
\our{}-Base & \bf 85.5 & \bf 53.4 \\
\bottomrule
\end{tabular}
\caption{Results on image classification and semantic segmentation.}
\label{tbl:unimodal_tasks}
\end{table*}

\begin{table*}[t]
\centering
\small
\begin{tabular}{lcccc}
\toprule
\multirow{2}{*}{\bf Stagewise Pre-Training} &  \multicolumn{2}{c}{\bf NLVR2} & \multicolumn{2}{c}{\bf Flickr30k} \\
 & dev & test-P & TR & IR \\
\midrule
Image-Only Pre-Training & 80.33 & 81.06 & 95.60 & 87.69 \\
Image-Only + Text-Only Pre-Training & \bf 82.09 & \bf 82.49 & \bf 95.67 & \bf 88.52 \\
\bottomrule
\end{tabular}
\caption{
Ablation studies of stagewise pre-training, i.e., different initialization for vision-language pre-training.
We report the average of R@1, R@5 and R@10 for Flickr30k.
Results of NLVR2 are averaged over three runs.
}
\label{tbl:ablation:init_models}
\end{table*}

\begin{table*}[t]
\centering
\small
\begin{tabular}{lccc|ccc|cccc}
\toprule
& \multicolumn{3}{c}{\textbf{Pre-Training Tasks}} &
\multicolumn{3}{c}{\textbf{Transformer}} &
\multicolumn{2}{c}{\textbf{NLVR2}} & \multicolumn{2}{c}{\textbf{Flickr30k}} \\
& ITC & ITM & MLM & Std TRM & MoME & MoME$-$VLExp & dev & test-P & TR & IR \\
\midrule
\tblidx{1} & \cmark & \xmark  &\xmark &\xmark & \cmark &\xmark & 58.51 & 58.83 & 92.23 & 84.24  \\
\tblidx{2} & \cmark & \xmark & \cmark & \xmark& \cmark &\xmark & 73.91 & 73.75 & 94.07 & 85.82  \\
\tblidx{3} & \cmark & \cmark & \xmark & \xmark& \cmark &\xmark & 76.46 & 76.19 & 94.37 & 85.67  \\
\tblidx{4} & \cmark & \cmark & \cmark & \cmark &\xmark & \xmark& 78.81 & 79.27 & 93.37 & 85.73  \\
\tblidx{5} & \cmark & \cmark & \cmark & \xmark&\xmark & \cmark & 79.58 & 80.11 & 94.50 & 86.69  \\
\tblidx{6} & \cmark & \cmark & \cmark & \xmark& \cmark & \xmark& \bf 80.13 & \bf 80.31 & \bf 95.17 & \bf 87.25  \\
\bottomrule
\end{tabular}
\caption{
Ablation studies of \mome{} Transformer and vision-language pre-training tasks. 
``ITC'' is short for image-text contrastive loss, ``ITM'' is image-text matching, and ``MLM'' is masked language modeling.
``Std TRM'' is short for standard Transformer, and ``MoME$-$VLExp'' is \mome{} without VL experts.
The average of R@1, R@5 and R@10 is reported for Flickr30k.
Results of NLVR2 are averaged over three runs.
}
\label{tbl:ablation:mome_tasks}
\end{table*}

\paragraph{Stagewise Pre-Training}

We first conduct ablation experiments of stagewise pre-training.
ViLT~\citep{vilt} shows that using the ViT~\citep{vit} model pretrained on image-only data as the initialization achieves better performance than the BERT model pretrained on text-only data.
Therefore we start experiments with image-only pre-training. 
We compare using image-only pre-training, and image-only pre-training plus text-only pre-training as the initialization.
For image-only pre-training, we directly use the parameters of \beit{}-Base to initialize the self-attention module and all modality experts.
For image-only pre-training plus text-only pre-training, we use pretrained parameters of \beit{}-Base to initialize the vision expert and self-attention module of \mome{} Transformer, and then pretrain its language expert on text corpora.
As shown in Table~\ref{tbl:ablation:init_models}, image-only pre-training plus text-only pre-training improves our vision-language model.
We also have tried to perform vision-language pre-training with random initialization but obtain a relatively low accuracy on downstream tasks.
Stagewise pre-training effectively leverages large-scale image-only and text-only corpus, and improves our vision-language pre-training.
Moreover, given the limited size of image-text pairs we used during pre-training, stage-wise pre-training on image-only and text-only data alleviates the need for image-text pair data.

\paragraph{\mome{} Transformer}

We also conduct ablation experiments of \mome{} Transformer.
We employ ViT-Base to initialize the models for the ablation experiments.
As present in Table~\ref{tbl:ablation:mome_tasks}, using \mome{} Transformer achieves better performance than standard Transformer for both retrieval and classification tasks.
In addition, we also analyse the contribution of vision-language expert (VL-FFN) used in \mome{} Transformer.
We remove the vision-language expert used in the top Transformer layers. 
Experimental results demonstrate that the introduction of vision-language expert improves the model.
Using vision-language expert captures more modality interaction.
Shared self-attention module used in \mome{} also positively contributes to our model.
Section~\ref{app:shared_att} presents the ablation study of shared self-attention module.

\paragraph{Pre-Training Tasks}

We perform ablation studies to analyse the contribution of different pre-training tasks, and the results are presented in Table~\ref{tbl:ablation:mome_tasks}.
Compared with the model trained only using image-text contrastive loss, our unified training performs much better across classification and retrieval tasks.
Introducing image-text matching with hard negative mining also greatly improves the model.
This demonstrates the effectiveness of our unified-training framework with \mome{} Transformer.
In addition, experimental results show that masked language modeling positively contribute to our model.
Please refer to the supplementary material for more ablation studies.

\paragraph{Global Hard Negative Mining}

Different from ALBEF~\cite{albef}, which samples hard negatives from training examples of the single GPU (named as local hard negative mining).
We perform hard negative mining from more candidates by gathering training examples of all GPUs (named as global hard negative mining).
As shown in Table~\ref{tbl:ablation:hard_neg}, 
our global hard negative mining brings significant improvements.

\begin{table*}[ht]
\centering
\small
\begin{tabular}{lcc}
\toprule
\multirow{2}{*}{\bf Models} &  \multicolumn{2}{c}{\bf NLVR2} \\
 & dev & test-P \\
\midrule
Local hard negative mining~\cite{albef} & 77.70 & 77.95 \\
Global hard negative mining (ours) & \bf 79.54 & \bf 79.48 \\
\bottomrule
\end{tabular}
\caption{
Global hard negative mining improves the model.
We perform experiments using 32 V100 GPUs. 
The batch size per GPU is 32, and the total batch size is 1024.
Local hard negative mining samples hard negatives from training examples of the single GPU (32 examples), while global hard negative mining uses training examples gathered from all GPUs as the candidates (1024 examples).
}
\label{tbl:ablation:hard_neg}
\end{table*}

\section{Conclusion}
In this work, we propose a unified vision-language pretrained model \our{}, which jointly learns a dual encoder and a fusion encoder with a shared \mome{} Transformer backbone. \mome{} introduces a pool of modality experts to encode modality-specific information, and aligns different modalities using the shared self-attention module. The unified pre-training with \mome{} enables the model to be used as a dual encoder for efficient vision-language retrieval, or as a fusion encoder to model cross-modal interactions for classification tasks. We also show that stagewise pre-training that leverages large-scale image-only and text-only corpus greatly improves vision-language pre-training. 
Experimental results demonstrate that \our{} outperforms previous state-of-the-art models on various vision-language classification and retrieval benchmarks. 

In the future, we would like to work on improving \our{} from the following perspectives:
\begin{itemize}
\item We will scale up the model size used in \our{} pre-training.
\item We are also interested in fine-tuning \our{} for vision-language generation tasks, such as image captioning, following the method proposed in UniLM~\citep{unilm}.
\item We are going to explore to what extent vision-language pre-training can help each other modality, especially as the shared \mome{} backbone naturally blends in text and image representations.
\item We can extend the proposed model to integrate more modalities (e.g., speech, video, and structured knowledge), supporting general-purpose multimodal pre-training.
\end{itemize}

\bibliographystyle{plainnat}
\bibliography{univlp}

\newpage
\appendix

\section{Ablation Study of Shared Self-Attention}
\label{app:shared_att}

Table~\ref{tbl:ablation:shared_attention} presents the ablation study of shared self-attention module used in \mome{} Transformer for encoding image patches and text tokens.
We compare shared self-attention with separate self-attention, which encodes image patches and text tokens using different attention parameters on the first L$-$F layers.
The shared self-attention used in \mome{} achieves better performance.
The shared self-attention module helps \our{} learn the alignment of different modalities, and fuse images and text at bottom layers for classification tasks.

\begin{table*}[ht]
\centering
\small
\begin{tabular}{lcccc}
\toprule
\multirow{2}{*}{\bf Transformer} &  \multicolumn{2}{c}{\bf NLVR2} & \multicolumn{2}{c}{\bf Flickr30k} \\
 & dev & test-P & TR & IR \\
\midrule
Separate Self-Attention & 78.92 & 78.95 & 94.63 & 86.88 \\
MoME (Shared Self-Attention) & \bf 80.13 & \bf 80.31 & \bf 95.17 & \bf 87.25 \\
\bottomrule
\end{tabular}
\caption{
Ablation study of the shared self-attention module used in \mome{}. We experiment with separate attention on the first L$-$F layers, which encodes image patches and text tokens using different attention parameters. 
}
\label{tbl:ablation:shared_attention}
\end{table*}

\section{Hyperparameters for Text-Only Pre-Training}
\label{app:text_pretraining}

For the text-only pre-training data, we use English Wikipedia and BookCorpus~\cite{bookcorpus}.
AdamW~\cite{adamw} optimizer with $\beta_1=0.9$, $\beta_2=0.98$ is used to train the models. 
The maximum sequence length is set to $196$.
The batch size is $1024$, and the peak learning rate is 2e-4.
We set the weight decay to $0.01$.
For the base-size model, we train the model for
$500$k steps.
The large-size model is trained for $200$k steps.

\section{Hyperparameters for Vision-Language Classification Fine-Tuning}
\label{app:finetune:cls}

\paragraph{Visual Question Answering (VQA)}
We fine-tune the models for $10$ epochs with $128$ batch size.
The peak learning rate is 3e-5 for the base-size model, and 1.5e-5 for the large-size model.
Following SimVLM~\cite{simvlm}, the input image resolution is $480 \times 480$. 
For \our{}-Large++, we use $768 \times 768$ image resolution.

\paragraph{Natural Language for Visual Reasoning (NLVR2)}

For results of Table 1, the models are fine-tuned for $10$ epochs with $128$ batch size.
The peak learning rate of the base-size and large-size models are set to 5e-5 and 3e-5, respectively.
The input image resolution is $384 \times 384$.
For ablation experiments, we fine-tune the models for $10$ epochs with $128$ batch size, and choose learning rates from \{5e-5, 1e-4\}.
The input image resolution is $224 \times 224$. All the ablation results of NLVR2 are averaged over $3$ runs.

\section{Hyperparameters for Vision-Language Retrieval Fine-Tuning}
\label{app:finetune:retrieval}

\paragraph{COCO}

We fine-tune the base-size model for $20$ epochs and large-size model for $10$ epochs with $2048$ batch size.
The peak learning rate is 2e-5 for the base-size model and 1e-5 for the large-size model.
The input image resolution is $384 \times 384$.

\paragraph{Flickr30K}

For results of Table 2, the base-size and large-size models are fine-tuned for $40$ epochs with a batch size of $2048$ and a peak learning rate of 1e-5. We use the fine-tuned model on COCO as the initialization. The input image resolution is $384 \times 384$.
For all ablation experiments, we fine-tune the models for $10$ epochs with $1024$ batch size.
The peak learning rate is set to 5e-5, and the input image resolution is $224 \times 224$.

\end{document}